\documentclass[runningheads]{llncs}

\usepackage{eccv}

\usepackage{eccvabbrv}

\usepackage{graphicx}
\usepackage{booktabs}
\usepackage{algorithm,algorithmicx,algpseudocode}
\usepackage{multirow}
\usepackage{makecell}

\usepackage[accsupp]{axessibility}  % Improves PDF readability for those with disabilities.

\usepackage{hyperref}

\usepackage{orcidlink}

\begin{document}

\title{Transferable 3D Adversarial Shape Completion using Diffusion Models} 

\titlerunning{Transferable 3D Adversarial Shape Completion}

\author{Xuelong Dai\inst{1}\orcidlink{0000-0001-6646-6514} \and
Bin Xiao\inst{1}\orcidlink{0000-0003-4223-8220}}

\authorrunning{X.~Dai et al.}

\institute{
The Hong Kong Polytechnic University\\
\email{xuelong.dai@connect.polyu.hk, b.xiao@polyu.edu.hk }}

\maketitle

\begin{abstract}
Recent studies that incorporate geometric features and transformers into 3D point cloud feature learning have significantly improved the performance of 3D deep-learning models. However, their robustness against adversarial attacks has not been thoroughly explored. Existing attack methods primarily focus on white-box scenarios and struggle to transfer to recently proposed 3D deep-learning models. Even worse, these attacks introduce perturbations to 3D coordinates, generating unrealistic adversarial examples and resulting in poor performance against 3D adversarial defenses. In this paper, we generate high-quality adversarial point clouds using diffusion models. By using partial points as prior knowledge, we generate realistic adversarial examples through shape completion with adversarial guidance. The proposed adversarial shape completion allows for a more reliable generation of adversarial point clouds. To enhance attack transferability, we delve into the characteristics of 3D point clouds and employ model uncertainty for better inference of model classification through random down-sampling of point clouds. We adopt ensemble adversarial guidance for improved transferability across different network architectures. To maintain the generation quality, we limit our adversarial guidance solely to the critical points of the point clouds by calculating saliency scores. Extensive experiments demonstrate that our proposed attacks outperform state-of-the-art adversarial attack methods against both black-box models and defenses. Our black-box attack establishes a new baseline for evaluating the robustness of various 3D point cloud classification models.

  \keywords{3D Black-box Adversarial Attacks \and Diffusion Models \and Model Uncertainty}
\end{abstract}

\section{Introduction}
Deep-learning models have demonstrated their overwhelming performance on 2D \cite{he2016deep,liu2021swin} and 3D computer vision \cite{Xiang_2021_ICCV,guo2021pct,ren2022benchmarking} tasks. An increasing number of applications rely on deep-learning models to achieve efficient and accurate services. Therefore, the security of deep-learning models is crucial and significant. 

Similar to the 2D scenario \cite{madry2017towards,carlini2017towards,croce2020reliable,liang2023styless,li2023physical}, 3D point cloud deep learning is also susceptible to adversarial attacks \cite{liu2019extending,xiang2019generating,zheng2019pointcloud}. These 3D adversarial attacks generate adversarial examples by introducing perturbations to the $xyz$ coordinates. However, such perturbations often lead to a significant degradation in visual quality, which can be easily detected by humans. Subsequent studies \cite{zhao2020isometry,wen2020geometry,Huang_2022_CVPR} have aimed to create less perceptible perturbations by taking into account geometric characteristics. Despite this, these attacks have been shown to perform poorly against defenses \cite{ji2023benchmarking}. Moreover, most existing attacks primarily focus on white-box settings, limiting their practicality in real-world scenarios. Existing black-box attacks \cite{hamdi2020advpc,he2023generating} mainly target early 3D point cloud deep-learning models, leaving a substantial gap in the learning between adversarial and benign models.

\begin{figure}[t]
   \begin{center}
     \includegraphics[width=1.0\linewidth]{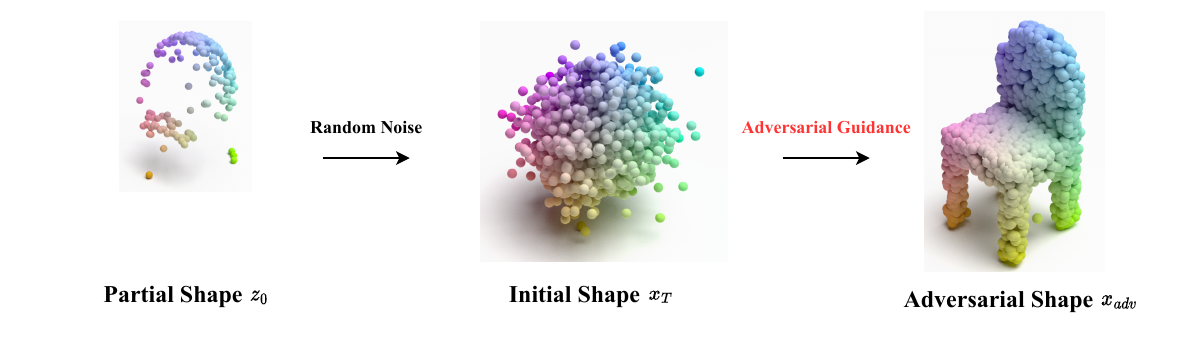}
   \end{center}
      \caption{\textbf{The adversarial shape completion.} Starting from the partial shape $z_0$, we construct our adversarial shape $x_{adv}$ by utilizing diffusion models with proposed adversarial guidance.}
   \label{fig:1}
   \end{figure}

In this paper, our objective is to execute high-quality black-box 3D adversarial attacks using diffusion models. To generate natural adversarial point clouds, we employ diffusion models, which are state-of-the-art generative models known for creating high-quality 2D images \cite{dhariwal2021diffusion,rombach2022high} and 3D point clouds \cite{zhou20213d,zeng2022lion}. It has been demonstrated that 2D diffusion models can generate adversarial examples \cite{chen2023advdiffuser,dai2023advdiff} by altering the diffusion process. By extension, it is intuitive that 3D diffusion models, with their impressive generation performance, are capable of creating adversarial examples. Specifically, we craft adversarial examples by employing diffusion models for shape completion tasks, as shown in Figure \ref{fig:1}. Using a partial shape as prior knowledge, our attack generates adversarial examples by completing shapes with the proposed adversarial guidance. Our approach to conducting adversarial attacks involves generating unseen data rather than introducing perturbations to clean data, effectively addressing the issue of unrealistic perturbations to $xyz$ coordinates.

In order to enhance the transferability of our crafted adversarial examples against black-box 3D models, we initially incorporate model uncertainty into the gradient inference of the substitute models. Li et al. \cite{li2023making} demonstrated that the introduction of probability measures to the substitute models can significantly enhance the performance of black-box attacks. They execute adversarial attacks by training the substitute model in a Bayesian manner. In our attack, we leverage the characteristics of 3D point clouds and incorporate model uncertainty through a Monte Carlo estimate over the inference from multiple down-sampled point clouds. Additionally, to improve the attack transferability against various network architectures, we employ ensemble logits to generate the adversarial guidance for the 3D diffusion model. To preserve the generation quality, we limit our adversarial guidance solely to the critical points that are selected based on the saliency scores. Our proposed black-box attack is capable of conducting black-box adversarial attacks against state-of-the-art 3D point cloud deep-learning models without the need to re-train the diffusion model.

Our contributions are summarized as follows:
\begin{itemize}
    \item We generate adversarial examples through shape completion using diffusion models, offering a novel perspective on the creation of imperceptible adversarial examples. The proposed attack introduces diffusion models to the topic of 3D adversarial robustness.
    \item We propose a variety of strategies to enhance the transferability of the proposed attacks without compromising the quality of generation. These strategies include: employing model uncertainty for improved inference of predictions, ensemble adversarial guidance to boost attack performance against unseen models, and generation quality augmentation to identify critical points and maintain the quality of generation.
    \item We conduct a comprehensive evaluation against existing state-of-the-art black-box 3D deep-learning models. Our experiments demonstrate that our proposed attack achieves state-of-the-art performance against both black-box models and defenses.
\end{itemize}

\section{Background}

\subsection{3D Point Cloud Classification}

The field of 3D point cloud classification poses unique challenges compared to 2D image classification, primarily due to the disorder and discrete nature of 3D point cloud data. Traditional 2D deep-learning models find it challenging to process such data efficiently. PointNet \cite{qi2017pointnet} stood out as the pioneering approach to address the challenge of 3D feature learning. PointNet largely enhanced the performance of 3D classification tasks by employing a symmetric function that effectively extracts features from the inherently disorderly input of 3D point cloud data. The success of PointNet \cite{qi2017pointnet} has sparked a surge in research focused on 3D deep learning. In an effort to enhance the performance of 3D feature learning, researchers have integrated graph convolutional operations to extract features from both local neighbors and the global shape of the point cloud. Two notable state-of-the-art 3D deep learning networks, PointNet++ \cite{qi2017pointnet++}  and DGCNN \cite{wang2019dynamic}, had successfully adopted graph convolutional layers. Recent approaches have further improved 3D point cloud classification by incorporating geometry features and transformers \cite{Xiang_2021_ICCV,guo2021pct,ren2022benchmarking}. These advancements contribute to achieving satisfying performance in the challenging task of 3D point cloud classification.

\subsection{3D Point Cloud Adversarial Attack}

3D deep-learning models exhibit vulnerability to adversarial attacks, even when using 2D adversarial approaches. However, the perturbations applied to 3D point cloud data are more perceptible to humans due to the specific data structure of point clouds. Adversarial perturbations that shift coordinates lead to noticeable changes in the original shape of 3D objects, presenting a challenge in devising stronger and more realistic adversarial attack methods. Early adversarial attack methods, such as those proposed by Liu et al.  \cite{liu2019extending} and Xiang et al.  \cite{xiang2019generating}, involve adding points generated from 2D FGSM, PGD, and C\&W attack methods. Zheng et al. \cite{zheng2019pointcloud}  demonstrated high attack performance on the PointNet network by dropping points with the lowest salience scores based on the saliency map. However, these attacks are easily detectable as they alter the number of points in the clean point cloud.

Subsequent works aim to create imperceptible perturbations by shifting point coordinates within the clean point clouds. Approaches like  ISO \cite{zhao2020isometry}, GeoA3 \cite{wen2020geometry}, SI-Adv \cite{Huang_2022_CVPR}, and PF-Attack \cite{he2023generating} achieved imperceptible shifting by leveraging geometric and shape information from clean point clouds.  LG-GAN \cite{zhou2020lg} and AdvPC \cite{hamdi2020advpc} utilized generative models to generate camouflaged perturbations effectively. However, only AdvPC and PF-Attack achieved effective black-box attacks against 3D point cloud classifiers. Nonetheless, these methods face challenges in being effective against recently proposed state-of-the-art 3D deep-learning models, resulting in a huge gap in the development between adversarial attacks and benign models.

\section{Preliminary}

\subsection{Threat Model}
Consider a point cloud $x \in \mathcal{P}^{K \times 3}$ consisting of $K$ points, where each point $x_i \in \mathcal{P}^3$ is represented by 3D $xyz$ coordinates. A classifier $f$ is employed to classify the input point cloud and assign a label, denoted as $f(x) \rightarrow y$. In the context of adversarial attacks, an adversary seeks to generate an adversarial example $x_{adv}$ with the objective of causing the target classifier $f$ to produce an incorrect classification result, represented as $y_{adv}$. Formally, the goal of the point cloud adversarial attack is defined as:

\begin{equation}
    \min D(x,x_{adv}),~~~~~~~~~ \text{s.t.} ~f(x_{adv})  = y_{adv}
    \label{eq:1}
\end{equation}
Equation \ref{eq:1} is designed to generate an imperceptible adversarial example $x_{adv}$ from the original point cloud $x$. This paper primarily concentrates on untargeted attacks, where $y_{adv}$ can be any label distinct from the ground truth label $y$.

\subsection{3D Point Cloud Generation and Completion}

Recent advancements in diffusion models \cite{ho2020denoising, dhariwal2021diffusion, rombach2022high, kim2022diffusionclip} applied to 2D image generation have showcased remarkable performance in terms of both generation quality and diversity. Likewise, recent studies on 3D diffusion models \cite{zhou20213d, luo2021diffusion, zeng2022lion} have demonstrated state-of-the-art performance in 3D point cloud generation tasks. The 3D denoising diffusion probabilistic model generates 3D point clouds with a denoising generation process. Starting from Gaussian noise $x_T$, the denoising process gradually produces the final output by a sequence of denoising-like steps, i.e., $x_T, x_{T-1}, \ldots, x_0$. 

The generative diffusion model, denoted as $p_\theta(x_{0:T})$, aims to learn the Gaussian
transitions from $p(x_T)=\mathcal{N}(x_T;0,\mathbf{I})$ by reconstructing $x_0$ from the diffusion data distribution $q(x_{0:T})$. This distribution introduces Gaussian noise to $x_0$ over the course of $T$ steps.  More specifically, these processes of adding noise and subsequent denoising can be formulated as a Markov transition:

\begin{equation}
\label{eq:2}
\begin{aligned}
    q(x_{0:T}) &= q(x_0)\prod_{t=1}^{T} q(x_t | x_{t-1})\\
p_\theta(x_{0:T}) &=  p(x_T)\prod_{t=1}^{T} p_\theta(x_{t-1} | x_t)
\end{aligned}
\end{equation}
where we name the $q(x_t | x_{t-1})$ as \textit{forward diffusion process} and $p_\theta(x_{t-1} | x_t)$ as \textit{reverse generative process}. Each detailed transition for each process is defined in accordance with the scheduling function $\beta_1, \dots, \beta_T$:

\begin{equation}\label{eq:e}
\begin{aligned}
q(x_t|x_{t-1})&:=\mathcal{N}(x_t:\sqrt[]{1-\beta_t}x_{t-1},\beta_t\textbf{I} )    \\
p_\theta(x_{t-1}|x_t) &:= \mathcal{N}(x_{t-1}:\mu_{\theta}(x_t,t),\sigma_t^2 \textbf{I})
\end{aligned}
\end{equation}
where $\mu_{\theta}(x_t,t)$ is the inference of the diffusion model to predict the shape of the point cloud. We set $\sigma_t^2 = \beta_t$ based on empirical knowledge.

The 3D point cloud generation task can be easily modified to achieve shape completion with an fixed partial shape $z_0 \in \mathcal{P}^{K_{p} \times 3}$ \cite{zhou20213d}. The \textit{forward diffusion process} and \textit{reverse generative process} are formulated as:

\begin{equation}\label{eq:4}
\begin{aligned}
q(\Tilde{x}_t|\Tilde{x}_{t-1},z_0)&:=\mathcal{N}(\Tilde{x}_t:\sqrt[]{1-\beta_t}\Tilde{x}_{t-1},\beta_t\textbf{I} )    \\
 p_\theta(\Tilde{x}_{t-1}|\Tilde{x}_t,z_0) &:= \mathcal{N}(\Tilde{x}_{t-1}:\mu_{\theta}(x_t,z_0,t),\sigma_t^2 \textbf{I})
\end{aligned}
\end{equation}

While recent studies have extensively explored the generation capabilities of 3D diffusion models, their potential in crafting adversarial point clouds remains largely unexplored. In this paper, we aim to generate high-quality adversarial point clouds with the reverse generative process of pre-trained 3D diffusion models. Note that we don't modify the training part of pre-trained models.
\section{Methodology}

\subsection{Diffusion Model for 3D Adversarial Shape Completion}

In crafting high-quality adversarial examples, our aim is to utilize diffusion models for their superior performance in 3D point cloud generation. Unlike previous generative models, the denoising generation process of diffusion models can naturally incorporate adversarial objectives \cite{chen2023advdiffuser,dai2023advdiff}, which can be viewed as a process of iterative adversarial attacks. Previous perturbation-based adversarial attacks perturb each point in the clean point cloud, commonly altering the shape of the original point cloud. In our work, we aim to minimize the impact of adversarial perturbations on the point cloud data and achieve adversarial attacks with our proposed method, the 3D adversarial shape completion attack.

The proposed attack generates adversarial point clouds with a fixed partial shape $z_0 \in \mathcal{P}^{K_{p} \times 3}$. We utilize any pre-trained 3D shape completion diffusion model $\epsilon_\theta$  to gradually generate the completed adversarial point cloud $x_0 = (z_0, \Tilde{x}_0)$ through the reverse generative process $p_\theta(\Tilde{x}_{t-1}|\Tilde{x}_t,z_0)$, $t = T,\ldots, 1$. For any intermediate shape $x_t = (z_0, \Tilde{x}_t)$, the adversarial generative process is defined as:

\begin{equation}
p_\theta(\Tilde{x}_{t-1}|\Tilde{x}_t,z_0) := \mathcal{N}(\Tilde{x}_{t-1}:\mu_{\theta}(x_t,z_0,t),\beta_t \textbf{I})- a\beta_t\nabla_{{{x}}_{t}} \mathcal{L}(f({{x}}_{t}), y)
    \label{eq:6}
\end{equation}
where $y$ represents the ground truth label of the original point cloud, $\mathcal{L}$ denotes the cross,  and the scale of adversarial guidance $a \in (0,1)$. We employ the untargeted I-FGSM-like gradient as the adversarial guidance for the adversarial generative process \cite{chen2023advdiffuser}.

We sample benign $\Tilde{x}_{t-1}$ from $\mathcal{N}(\Tilde{x}_{t-1}:\mu_{\theta}(x_t,z_0,t),\beta_t \textbf{I})$ by following PVD \cite{zhou20213d}:
\begin{equation}
\label{eq:pvd}
    \Tilde{x}_{t-1} = \frac{1}{\sqrt{\alpha_t}}\left(\Tilde{x}_t - \frac{1 - \alpha_t}{\sqrt{1-\tilde{\alpha}_t}} \epsilon_\theta(\Tilde{x}_t, z_0, t)\right) + \sqrt{\beta_t} \varepsilon,
\end{equation}
where $\alpha$ and $\beta$ are hyper-parameters from the pre-trained $\epsilon_\theta$, and $\varepsilon \sim N(0, \textbf{I})$. 

%By following the adversarial generative process, we will generate the 3D adversarial examples $x_{adv}$ with the unchanged partial shape $z_0$ against the white-box classifier $f$ with any pre-trained 3D shape completion diffusion model. 

\subsection{Diffusion Model with Boosting Transferbility}

In order to improve the effectiveness of the proposed attack on a black-box target model, we have outlined several effective strategies to enhance the transferability of the generated 3D point clouds, all without increasing the magnitude of the adversarial guidance.

\noindent \textbf{Employing Model Uncertainty}. Previous works \cite{li2019generative,carbone2020robustness} have shown that leveraging model uncertainty for feature learning is proposed to be more robust to adversarial attacks compared to standard deep learning models. These Bayesian deep neural networks are probabilistic models that predict input by computing expectations from maximum likelihood estimation over model parameters. Furthermore, utilizing model uncertainty \cite{li2023making} demonstrates improved adversarial transferability. However, the application of model uncertainty in 3D contexts is currently underexplored. Considering the characteristics of 3D point clouds, which comprise unordered 3D points, the removal of some points does not alter the classification outcome of the original point cloud \cite{zhou2019dup}. Therefore, we are able to straightforwardly adopt model uncertainty to 3D deep-learning models with the \textit{MC dropout}-like \cite{gal2016dropout} approach over the input. In our attack, we adopt Simple Random Sampling over the 3D point clouds and use the Monte Carlo estimate over $M$ re-sampled point clouds to obtain the estimated adversarial guidance:
\begin{equation}
    \nabla_{{{x}}_{t}}\mathcal{L}_{\text{MU}}(f({{x}}_{t}), y)) =\frac{1}{M}\sum_{s=1}^{M}\nabla_{{{x}}_{s}}\mathcal{L}(f({{x}}_{s}),y)
    \label{eq:7}
\end{equation}
The ${{x}}_{s}$ is obtained by simple random sampling from ${{x}}_t= (z_0, \Tilde{{x}}_{t})$:
\begin{equation}
    P_i(\Tilde{{x}}_t)=\{\mathrm{1}_x|x\in\Tilde{{x}}_t, \mathrm{1}_x\sim Ber(0.5)\}
    \label{eq:8}
\end{equation}
where $x$ is sampled from a $Bernoulli(0.5)$ distribution to indicate the existence of $x$ in the ${{x}}_{s} = (z_0, \Tilde{{x}}_{s})$ point cloud re-sampled from $i^\text{th}$ point of $\Tilde{x}_{t}$, and $z_0$ is not re-sampled.

\noindent \textbf{Ensemble Adversarial Guidance}. In the 2D attack scenario, the ensemble attack is an effective way to enhance the attack transferability by utilizing multiple white-box models to calculate the average gradient of the objective loss. Ensemble gradient in 2D results in perturbation in the given pixel of the 2D image. In our attack, we ensemble the logits of selected substitute models according to the generative process in Equation \ref{eq:6}. Formally, with $n_\text{ens}$ substitute models, the ensemble adversarial objective function is defined as:
\begin{equation}
    \mathcal{L}(f_{{ens}}({{x}}_{t}),y)=-\log(\text{softmax}\sum_{n=1}^{n_\text{ens}}w_{n}p_{f_{n}}(y|{{x}}_{t}))   
    \label{eq:9}
\end{equation}
where $w_n$ is the weight parameters, and we use the proportion of correctly classified point clouds for an adaptive ensemble attack; $p_f$ is the predictive distribution of $f$.

\noindent \textbf{Generation Quality Augmentation}. Previous work \cite{zheng2019pointcloud} has shown that individual points within a point cloud can have varying degrees of impact on the classification outcome of a 3D deep-learning model. This insight suggests that identifying critical points within the point cloud could achieve strong adversarial attacks. Due to the significant reduction in visual quality caused by perturbations to 3D coordinates, it is advisable to control these perturbations by constraining the $\ell_0$ distance between the adversarial and benign point clouds. Thus, our objective is to create adversarial examples by altering only a subset of $N$ points of the benign point cloud. The saliency score of given point $x$ is calculated as:
\begin{equation}
    \text{score}_{x}= \sum_{3}\frac{\partial \mathcal{L}(f({{x}}_{t}),y)  }{\partial x } 
    \label{eq:10}
\end{equation}
where the saliency score is the sum of $xyz$ channels of point $x$. Moreover, we further adopt $\ell_\text{inf}$ norm restriction to the perturbation at each diffusion step for a fair comparison with perturbation-based adversarial attacks.

\subsection{Transferable 3D Adversarial Shape Completion Attack}

We summarize the proposed black-box 3D adversarial attack in Algorithm \ref{alg:1}. In the early generation process, the generated point clouds are disorganized. Therefore, we only perform adversarial guidance at given timestep $T_\text{adv}$. We apply the $\text{Clip}$ \cite{goodfellow2014explaining} function to the $\ell_\text{inf}$ norm to limit the perturbation in adversarial guidance.

\begin{algorithm}[tb]

  \caption{Transferable 3D Adversarial Shape Completion Attack Algorithm} 
  \label{alg:1}
  
  \begin{algorithmic}[1]
    \Require $f_\text{ens}$: substitute models
    \Require $z_0$: partial shape for shape completion
    \Require $y$: class label for shape completion
    \Require $T$: reverse generation process timestep for LDM
    \Require $T_\text{adv}$: timestep for adversarial guidance
    \Require $N$: number of perturbed points at each diffusion step
    \Require $M$: number of simple random sampling 

    \State $\Tilde{x}_T \sim \mathcal{N}(0, \textbf{I})$, ${{x}}_{T}=(z_0, \Tilde{x}_T)$
    \State $x_{adv} = \varnothing $
    \For{$t=T, \dotsc, 1$}
      \If{$t$ is in $T_\text{adv}$} 
      \State Sample $\Tilde{x}_{t-1}$ with Equation \ref{eq:4}
      \For{$m=1,\dotsc, M$}
      \State Simple random sampling with Equation \ref{eq:8}
      \State Obtain the ensemble adversarial loss with Equation \ref{eq:9}
      \EndFor
      \State  Monte Carlo estimate with Equation \ref{eq:7}
      \State  Calculate the saliency score of $\Tilde{x}_{t-1}$ with Equation \ref{eq:10}
      \State  Update top-$N$ points from step 11 of $\Tilde{x}_{t-1}$ with Equation \ref{eq:6}
      \State  $\Tilde{x}_{t-1} = \text{Clip}(\Tilde{x}_{t-1})$
      \Else
      \State Sample $\Tilde{x}_{t-1}$ with Equation \ref{eq:4}
      \EndIf
    \EndFor
      \State $x_0 = (z_0, \Tilde{x}_0)$
      \State  $x_{adv} \gets x_{0}$ if $f_\text{ens}(x_{0})\neq y$
    
    \State \textbf{return} $x_{adv}$
  \end{algorithmic}
\end{algorithm}

\subsection{Revisiting 3D Black-Box Adversarial Attack}

%Black-box adversarial attack is a much more challenging attack than the white-box adversarial attack, even worse, 3D black-box adversarial attack is harder than 2D scenario. As shown in Figure ASDASD, the data distribution of the existing ShapeNet 3D dataset is long-tailed. Therefore, existing adversarial attack methods achieve relatively higher ASR on classes with less data. This circumstance is even worse in the ModelNet40 dataset with only 10000 samples in 40 classes. Therefore, we use the ShapeNet dataset rather than ModelNet40 in this paper. Another challenging problem on the 3D black-box adversarial attack is the different model architectures. To provide a detailed discussion of the transferability between different 3D models, we demonstrate the cosine similarity of various models in FIGUREW sASF. The result shows that the gradients from models with different model architectures significantly differ,  thereby presenting a substantial challenge for 3D black-box adversarial attacks. These challenging problems make existing 3D black-box adversarial attacks only effective against several 3D models on ModelNet40 dataset.

\begin{figure}[t]
   \begin{center}
     \includegraphics[width=1.0\linewidth]{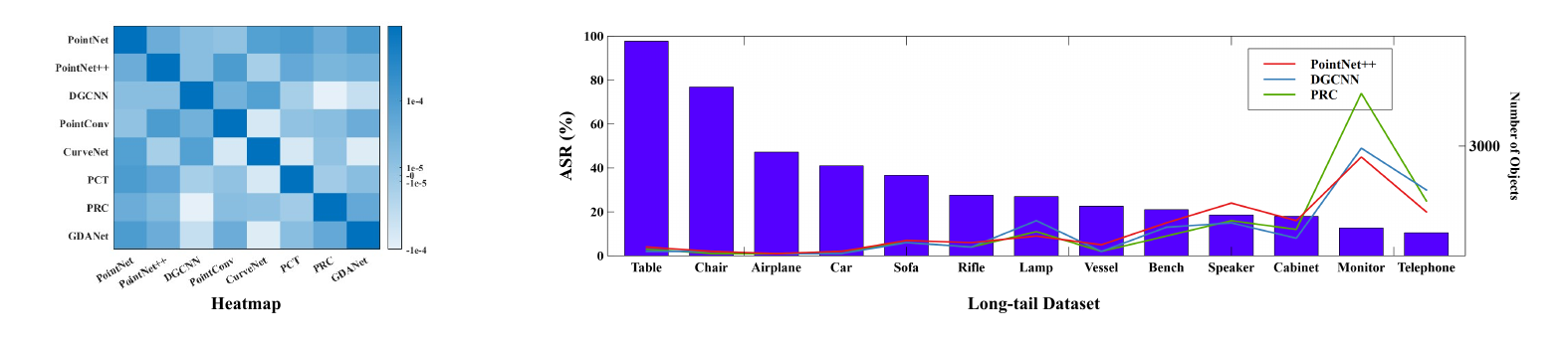}
   \end{center}
      \caption{\textbf{The challenging 3D black-box adversarial attacks.} The value in the Heatmap is re-scaled for better visualization. We use the top 13 classes from the ShapeNet dataset to demonstrate the long-tailed dataset problem. We use PGD with $\ell_\text{inf}=0.16$ on PointNet to evaluate the black-box attack success rate (ASR).}
   \label{fig:3d}
   \end{figure}

Black-box adversarial attacks present a significantly greater challenge than white-box adversarial attacks, with 3D black-box adversarial attacks proving even more difficult than their 2D counterparts. As illustrated in Figure \ref{fig:3d}, the data distribution of the existing ShapeNet 3D dataset is long-tailed. Consequently, existing adversarial attack methods tend to achieve a higher ASR on classes with less data (the top 5 classes contain 50\% data but only contribute 14\% success adversarial examples). This issue is similar in the ModelNet40 dataset, in which the top 5 classes contain 30\% of data. Another significant challenge in 3D black-box adversarial attacks lies in the varying model architectures. To provide a comprehensive discussion on the transferability between different 3D models, we have demonstrated the cosine similarity of various models in Figure \ref{fig:3d}. The results indicate that gradients from models with different architectures vary significantly, thus posing a considerable challenge for 3D black-box adversarial attacks. These challenging problems make existing 3D black-box adversarial attacks effective against only a few 3D models on the ModelNet40 dataset.

To execute an effective black-box 3D adversarial attack, we employ diffusion models to directly generate adversarial examples. The gradual diffusion generation process allows for the introduction of adversarial guidance with significantly less perturbation than existing adversarial attacks. Adversarial shape completion aids in identifying the vulnerable rotation for more potent adversarial attacks and ensures the reliable generation of natural point clouds, surpassing shape generation tasks. In addition to utilizing an ensemble attack approach, we also employ random sampling to leverage model uncertainty and enhance performance against defenses. By taking into account the characteristics of 3D point clouds and the generation performance of diffusion models, we are able to achieve an effective and high-quality black-box 3D adversarial attack.

\begin{table}[t]
\caption{\textbf{The attack success rate (ASR \%) of transfer attack on the ShapeNet dataset.} The adversarial examples of existing attack methods are generated from the PointNet model. The Average ASR is calculated among the seven black-box models (3DAdvDiff$_\text{ens}$ is calculated among the five black-box models). }

\vspace{-0.2in}
\begin{center}
\resizebox{0.99\columnwidth}{!}{  
\begin{tabular}{c|l|c|cccccccc}

\Xhline{3\arrayrulewidth}
Dataset                                     & Method    & PointNet & PointNet++ & DGCNN & PointConv & CurveNet & PCT & PRC & GDANet & Average \\ \hline
\multicolumn{1}{c|}{\multirow{7}{*}{Chair}} & PGD       &   99.7       &    1.0        &   0.9    &   1.2        &    0.7      &  1.4   &   0.9  &   2.1     &   1.2      \\
 & KNN       &   99.2       &    0.8        &   0.8    &   1.0        &    0.4      &  1.2   &   1.0  &   2.1     &   1.0      \\
& GeoA3     &  99.6        &    0.9        & 0.8      &  1.2         &  0.7        &  0.8   &  1.0   &  0.9      &   0.9      \\
& SI-Adv    &  82.4        &   1.2         &  1.2     &    1.5       &  1.5        &   1.4  &   2.3  &    2.2    &    1.6     \\ \cline{2-11} 
& AdvPC     &   71.8       &     2.2       &   0.9    &   1.5        &    1.8      &  2.1   &   2.6  &   2.0     &    1.6     \\
& PF-Attack &    99.0      &    20.2        &  5.6     &   4.8        &   3.2       &   1.0  &  2.5   &   1.6     &    5.5     \\
& 3DAdvDiff    &    99.9      &    60.6        &   8.7     &   23.5        &     9.8     &  6.9   &  14.9   &   8.9     &     19.0    \\
& 3DAdvDiff$_\text{ens}$    &   99.9       &    \textbf{94.5}        &    \textbf{\textit{99.9}}   &     \textbf{91.3}      &    \textbf{88.6}      & \textbf{65.8}   &  \textbf{\textit{99.9}}    &  \textbf{85.6}      &    \textbf{85.2}     \\ 
\Xhline{3\arrayrulewidth}
\multicolumn{1}{c|}{Dataset}                & Method    & PointNet & PointNet++ & DGCNN & PointConv & CurveNet & PCT & PRC & GDANet & Average \\ \hline
\multicolumn{1}{c|}{\multirow{7}{*}{All}}   & PGD       &   99.9       &   2.1        &   0.7    &   0.8        &   0.5      &  0.4   &   0.7  &   1.6   &   0.9     \\
 & KNN       &   99.9       &   2.2        &   0.7    &   0.7        &    0.5      &  0.6   &   1.1  &   1.6     &   1.1      \\
& GeoA3     &  99.8        &    2.0        & 1.5      &  1.4         &  0.9        &  0.6   &  0.9   &  1.1      &   1.2      \\
& SI-Adv    &  92.5        &   2.0         &  1.7     &    1.5       &  1.2        &   1.0  &   1.3  &    1.0    &    1.4     \\ \cline{2-11} 
& AdvPC     &   89.6       &     0.4       &   0.2    &   0.5        &    0.4      &  0.6   &   0.7  &   0.5     &    0.5     \\
& PF-Attack &    99.6      &    24.2        &  6.7     &   5.1        &   3.8       &   1.2  &  2.4   &   1.9     &    6.2     \\
& 3DAdvDiff    &    99.9      &    73.2        &   12.6     &   55.3        &     40.5    &  32.6   &  25.9   &   16.0     &     36.6   \\
& 3DAdvDiff$_\text{ens}$    &      99.9    &    \textbf{97.0}        &   \textbf{\textit{99.9}}     &    \textbf{94.5}       &  \textbf{93.5}        & \textbf{80.5}   &  \textbf{\textit{99.9}}    &  \textbf{85.2}      &    \textbf{90.1} \\ 
\Xhline{3\arrayrulewidth}
\end{tabular}
}
\end{center}
   
\vspace{-0.2in}
\label{tab:1}
\end{table}

\section{Experiments}

\subsection{Experimental Setup}
\noindent \textbf{Dataset}. Due to ModelNet40 being insufficient to train the diffusion model, we use the ShapeNet \cite{chang2015shapenet} dataset for major evaluations. The ShapeNetCore split is adopted, which contains 55 categories with 42003 data, of which 31535 samples are used for training and 10468 samples are used for testing. We select PVD \cite{zhou20213d} for the diffusion model in this paper. The proposed attack does not require additional training in the diffusion model, we follow settings as in the original PVD paper for selecting shape completion's partial shapes. Public checkpoints \cite{zhou20213d} from Airplane, Chair, and Car are selected for repeatability. Experiments on ModelNet40 are discussed in the Appendix.

\noindent \textbf{Target Models}. For a better evaluation of different network architectures, we select eight widely adopted 3D deep-learning models as the black-box models, including PointNet \cite{qi2017pointnet}, Pointnet++ (SSG) \cite{qi2017pointnet++}, DGCNN \cite{wang2019dynamic}, PointConv (SSG) \cite{wu2019pointconv}, CurveNet \cite{Xiang_2021_ICCV}, PCT \cite{guo2021pct}, PRC \cite{ren2022benchmarking}, and GDANet \cite{xu2021learning}.

\noindent \textbf{Comparisons}. We have chosen four white-box 3D adversarial attacks as our baseline for comparison, namely: PGD \cite{liu2019extending}, KNN \cite{tsai2020robust}, GeoA3 \cite{wen2020geometry}, and SI-Adv \cite{Huang_2022_CVPR}. We also employ existing black-box 3D adversarial attacks, specifically: AdvPC \cite{hamdi2020advpc} and PF-Attack \cite{he2023generating}. We use PointNet as the substitute model by default and the perturbations are constrained under the $\ell_\text{inf}$-normal ball with a radius of 0.16. We use 3DAdvDiff to denote the white-box version of the proposed attack and 3DAdvDiff$_\text{ens}$ for boosting transferability version.

\noindent \textbf{Defenses}. We select SRS \cite{zhou2019dup}, SOR \cite{zhou2019dup}, DUP-Net \cite{zhou2019dup}, IF-Defense \cite{wu2020if}, and Adversarial Hybrid Training \cite{ji2023benchmarking} for evaluation under defenses. All the defense settings are followed according to \cite{ji2023benchmarking}.

\noindent \textbf{Attack Settings}. We select PointNet, DGCNN, and PRC for ensemble adversarial guidance on 3DAdvDiff$_\text{ens}$. The hyper-parameters of the proposed attack are set to: $a=0.4, T=1000, T_\text{adv}=(0,0.2T],N=200,M=5,K=2048$. We also adopt $\ell_\text{inf}=0.16$ restriction to the adversarial guidance. We set 200 points for partial shapes. For each partial shape, we generate 20 views and only save the views that successfully attack the substitute models. To evaluate the attack performance, we use the top-1 accuracy of the target model to evaluate the Attack Success Rate (ASR). The experiment results are averaged over 10 attacks.

\subsection{Attack Performance}

\noindent \textbf{Transfer Attack}. We evaluate the transfer attack performance of current point cloud adversarial attack methods on selected robust classes. The results are given in Table \ref{tab:1}. As we discussed in Section 4.4, the adversarial examples from state-of-the-art attacks merely transfer to different models, particularly those recently developed 3D models. Models trained on long-tailed datasets typically exhibit limited generalization. However, our proposed white-box 3DAdvDiff achieves notably better performance even on the black-box adversarial attack. Furthermore, 3DAdvDiff$_\text{ens}$ considerably boosts the attack performance of 3DAdvDiff without augmenting the magnitude of the adversarial guidance, thereby validating the effectiveness of our proposed methods.

\begin{table}[t]
\caption{\textbf{The attack success rate (ASR \%) of different adversarial attack methods against defenses.} All attacks are evaluated under white-box settings against the PointNet model.}

\vspace{-0.2in}
\begin{center}
%\resizebox{1.0\columnwidth}{!}{  
\begin{tabular}{l|c|ccccc}

\Xhline{3\arrayrulewidth}
Method    & ASR & SRS & SOR & DUP-Net & IF-Defense & HybridTraining \\ \hline
PGD       & 99.9 &  5.9   & 1.0  &   0.7      &     13.8       &    1.9            \\
KNN       & 99.9 &  4.0   & 0.9  &   0.4      &     13.0       &    1.3    \\
GeoA3     & 99.8 & 4.9  &  1.6  &      0.8      &      13.6      &  2.2              \\
SI-Adv    & 92.5 &  10.8   & 0.9    &    0.9     &   14.9         &     2.0           \\ \hline
AdvPC     & 89.6 & 4.1    &  1.5   &   0.7      &    13.2        &      1.9          \\
PF-Attack & 99.6 &  8.5   &  3.6   &    2.8     &    13.9        &    2.0            \\
3DAdvDiff    & 99.9    & 82.2    &   9.9      &     9.6       &    \textbf{30.0}     &   9.4   \\
3DAdvDiff$_\text{ens}$    & 99.9 &  \textbf{85.9}   &  \textbf{49.1}   &    \textbf{36.9}     &   22.5         &  \textbf{96.1}     \\ 
\Xhline{3\arrayrulewidth}        
\end{tabular}
%}
\end{center}
   
\vspace{-0.2in}
\label{tab:2}
\end{table}

\noindent \textbf{Adversarial Defenses}. We evaluate the adversarial examples against a variety of defenses under white-box settings, as shown in Table \ref{tab:2}. The findings indicate that current defenses can effectively counter existing adversarial attacks, even with simple SRS (Simple Random Sampling). Defense methods that rely on outlier point removal exhibit the best performance among all defenses, suggesting that perturbation-based attack methods tend to displace points outside the original shape by adding perturbations to $xyz$ coordinates. Our proposed 3DAdvDiff significantly outperforms state-of-the-art adversarial attacks. Due to its utilization of model uncertainty, 3DAdvDiff is particularly effective against random sampling. The proposed critical point selection of 3DAdvDiff$_\text{ens}$ is effective against outlier removal defenses. However, the performance of 3DAdvDiff$_\text{ens}$ against IF-Defense is not satisfying due to the selection of critical points. Balancing generation quality and defense performance remains a challenge. In future work, we aim to enhance attack performance against reconstruction-based defenses.

\begin{figure}[t]
   \begin{center}
     \includegraphics[width=0.7\linewidth]{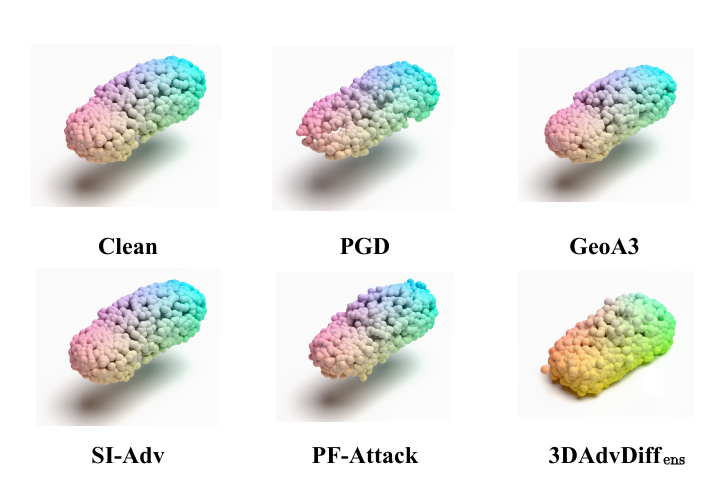}
   \end{center}
      \caption{\textbf{The visual quality of adversarial examples.} The black-box adversarial examples are relatively unnatural compared to white-box adversarial examples.}
   \label{fig:2}
   \end{figure}

\noindent \textbf{Generation Quality}. We further assess the distance between benign and adversarial examples to evaluate the visual quality of existing adversarial attack methods, as shown in Table \ref{tab:more}. The Chamfer Distance (CD), Hausdorff Distance (HD), and Mean Square Error (MSE) are selected. Given that we apply the same $\ell_\text{inf}=0.16$ norm to limit the perturbation for each attack, the visual quality across different attack methods is relatively similar. However, it is hard to give a fair comparison with 3DAdvDiff's adversarial examples, because the adversarial sampling of diffusion models can lead to the generation of new point clouds with completely different shapes. Therefore, the generation quality of 3DAdvDiff$_\text{ens}$ is evaluated by the difference between the benign samples and the adversarial examples with fixed sampling. A visual comparison is provided in Figure \ref{fig:2} for a more comprehensive demonstration. The point clouds generated by 3DAdvDiff$_\text{ens}$ is smoother than existing attacks.

\begin{table}[ht]
\caption{\textbf{The generation quality on the ShapeNet dataset.} The CD distance is multiplied by 10$^{-2}$. }

\vspace{-0.2in}
\begin{center}
%\resizebox{1.0\columnwidth}{!}{  

\begin{tabular}{l|ccccccc}

\Xhline{3\arrayrulewidth}
Method & \multicolumn{1}{l}{PGD} & \multicolumn{1}{l}{KNN} & \multicolumn{1}{l}{GeoA3} & \multicolumn{1}{l}{SI-Adv} & \multicolumn{1}{l}{AdvPC} & PF-Attack & 3DAdvDiff$_\text{ens}$ \\ \hline
HD     &          0.136               &     0.105                    &           0.039                &        0.071                    &      \textbf{0.028}                     &    0.046       &   \textit{0.098}      \\
CD     &         0.46                &    0.42                     &        \textbf{0.10}                   &         0.33                   &     0.27                      &    0.25       &   \textit{0.14}   \\
MSE     &       2.71                  &      2.42                   &          1.50                 &         3.08                   &      2.04                     &   1.85        &       \textbf{\textit{1.18}}
\\ \Xhline{3\arrayrulewidth} 
\end{tabular}
%}
\end{center}
   
\vspace{-0.2in}
\label{tab:more}
\end{table}

\noindent \textbf{Time efficiency}. Despite the proposed 3DAdvDiff achieves overwhelmingly performance on black-box adversarial attacks. The generation speed of diffusion models is a critical problem to influence its development. As shown in Table \ref{tab:time}, the running time of the proposed 3DAdvDiff is relatively slower than previous perturbation-based attack methods. However, we can improve the sampling speed by adopting DDIM sampling to PVD. Detailed discussion is given in the Appendix.

\begin{table}[ht]
\caption{\textbf{The average running time to generate one adversarial example.}}

\vspace{-0.2in}
\begin{center}
%\resizebox{1.0\columnwidth}{!}{  

\begin{tabular}{l|ccccccc}

\Xhline{3\arrayrulewidth}
Method & \multicolumn{1}{l}{PGD} & \multicolumn{1}{l}{KNN} & \multicolumn{1}{l}{GeoA3} & \multicolumn{1}{l}{SI-Adv} & \multicolumn{1}{l}{AdvPC} & PF-Attack & 3DAdvDiff$_\text{ens}$ \\ \hline
Time (s)     &          1.1               &     17.3                    &          81.6               &        7.0                    &      2.5                    &    38.6       &  60.8      \\
\Xhline{3\arrayrulewidth} 
\end{tabular}
%}
\end{center}
   
\vspace{-0.2in}
\label{tab:time}
\end{table}

\noindent \textbf{Integration with other methods}. To completely demonstrate the effectiveness of the proposed transferability boosting methods, we integrate the proposed improvement methods with existing attacks. As shown in Table \ref{tab:3}, our proposed enhancement methods markedly improve the performance of PGD, SI-Adv, and AdvPC on black-box attacks. However, the performance increase of adversarial attacks is limited without the diffusion models.

\begin{table}[ht]
\caption{\textbf{The ensemble of proposed boosting transferability methods with existing attack methods.} The experiments are performed on the \textbf{whole} test dataset of the ShapeNet dataset.}

\vspace{-0.2in}
\begin{center}
\resizebox{0.99\columnwidth}{!}{  
\begin{tabular}{l|c|cccccccc}

\Xhline{3\arrayrulewidth}
Method    & PointNet & PointNet++ & DGCNN & PointConv & CurveNet & PCT & PRC & GDANet & Average \\ \hline
PGD       &   \textbf{99.8}       &    10.8        &   8.9    &   11.1        &   7.1      &  7.3   &   9.1  &   10.1     &   9.2      \\
PGD + 3DAdvDiff       &   99.5       &    \textbf{48.9}        &   \textbf{\textit{93.6}}    &   \textbf{21.7}        &    \textbf{25.6}      &  \textbf{14.2}   &   \textbf{\textit{96.1}}  &   \textbf{14.5}     &   \textbf{25.0}      \\ \hline
SI-Adv    &  \textbf{97.6}        &   12.2         &  10.2     &    11.9       &  7.5        &   8.8  &  12.8  &    8.3    &    10.2     \\
SI-Adv + 3DAdvDiff     &   70.5       &     \textbf{42.8}       &   \textbf{\textit{45.9}}    &   \textbf{19.2}        &    \textbf{24.9}      & \textbf{20.4}   &  \textbf{\textit{38.6}}  &   \textbf{21.7}     &    \textbf{25.8}    \\ \hline
AdvPC     &  \textbf{96.9}        &    7.7       & 6.1      &  6.3         &  10.9        &  5.4   &  6.8   &  6.1      &  7.0      \\
AdvPC + 3DAdvDiff     &  95.2        &    \textbf{57.5}        & \textbf{\textit{75.8}}      & \textbf{38.1}         &  \textbf{35.4}        &  \textbf{21.8}   &  \textbf{\textit{63.0}}   & \textbf{16.1}      &   \textbf{33.8}      \\ 

\Xhline{3\arrayrulewidth}       
\end{tabular}
}
\end{center}
   
\vspace{-0.2in}
\label{tab:3}
\end{table}

\subsection{Ablation Study}

\begin{figure}[t]
   \begin{center}
     \includegraphics[width=1.0\linewidth]{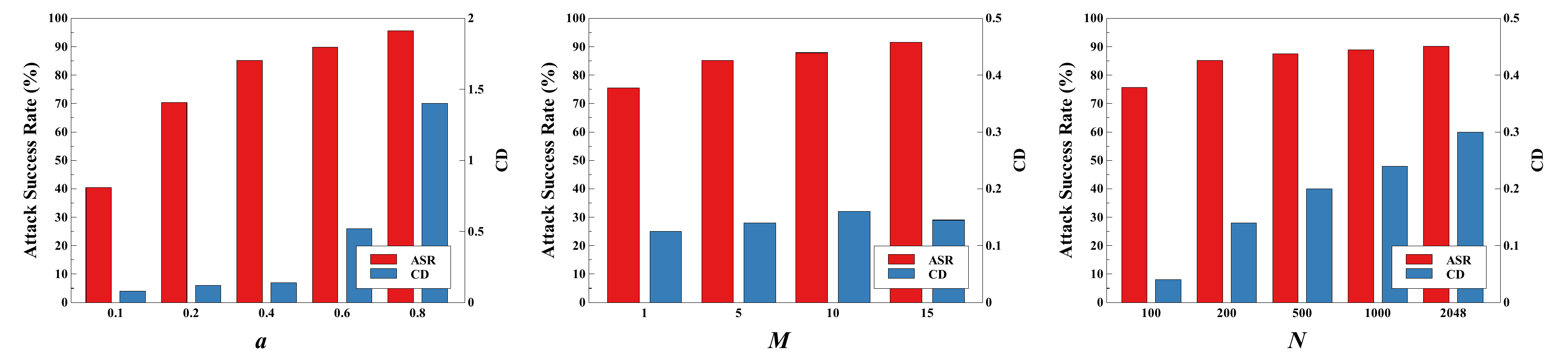}
   \end{center}
      \caption{\textbf{The ablation study of proposed 3DAdvDiff$_\text{ens}$.} The results are evaluated on the Chair class of the ShapeNet dataset. We use average ASR to test the black-box attack performance.}
   \label{fig:ab}
   \end{figure}
   
We conduct a series of ablation studies to investigate the effectiveness of various approaches in 3DAdvDiff$_\text{ens}$ for enhancing transferability, including model uncertainty, ensemble adversarial guidance, and generation quality augmentation.

\noindent \textbf{Adversarial Guidance}. The parameter $a$ of the adversarial guidance is critical to the attack success rate and the generation quality, as shown in Figure \ref{fig:ab}. However, our proposed 3DAdvDiff generates adversarial examples by finding the most vulnerable rotation from multiple views. Therefore, we can easily balance ASR and the generation quality without largely decreasing ASR.

\noindent \textbf{Model Uncertainty}. We evaluate the performance of model uncertainty with varying settings of $M$. Figure \ref{fig:ab} indicates that attack transferability increases with a larger $M$. However, this significantly impacts the time efficiency required to generate adversarial examples. As shown in Table \ref{tab:4}, incorporating model uncertainty significantly improves the transfer attack performance of 3DAdvDiff combined with the sampling of the diffusion model. These results further validate the effectiveness of our proposed model uncertainty approach.

\begin{table}[ht]
\caption{\textbf{The ensemble of model uncertainty with 3DAdvDiff.} The experiments are performed on the Chair class of the ShapeNet dataset.}

\vspace{-0.2in}
\begin{center}
\resizebox{0.99\columnwidth}{!}{  
\begin{tabular}{l|c|cccccccc}

\Xhline{3\arrayrulewidth}
Method    & PointNet & PointNet++ & DGCNN & PointConv & CurveNet & PCT & PRC & GDANet & Average \\ \hline
3DAdvDiff     &    99.9      &    60.6        &   8.7     &   23.5        &     9.8     &  6.9   &  14.9   &   8.9     &     19.0   \\
3DAdvDiff + MU      &  99.9        &    \textbf{82.6}        & \textbf{78.6}      &  \textbf{85.6}         &  \textbf{84.2}        &  \textbf{68.1}  & \textbf{59.5}   &  \textbf{70.2}      &   \textbf{75.5}     \\ 

\Xhline{3\arrayrulewidth}       
\end{tabular}
}
\end{center}
   
\vspace{-0.2in}
\label{tab:4}
\end{table}

\noindent \textbf{Ensemble Adversarial Guidance}. We test the performance of 3DAdvDiff with ensemble adversarial guidance. Table \ref{tab:ag} shows that the proposed adversarial guidance can effectively improve the performance of transfer attacks against black-box models. Simultaneously, the use of ensemble adversarial guidance does not compromise the generation quality of the proposed attack.

\begin{table}[ht]
\caption{\textbf{The performance of ensemble adversarial guidance.} The experiments are performed on the Chair class of the ShapeNet dataset.}

\vspace{-0.2in}
\begin{center}
\resizebox{0.99\columnwidth}{!}{  
\begin{tabular}{l|c|cccccccc}

\Xhline{3\arrayrulewidth}
Method    & PointNet & PointNet++ & DGCNN & PointConv & CurveNet & PCT & PRC & GDANet & Average \\ \hline
3DAdvDiff     &    99.9      &    60.6        &   8.7     &   23.5        &     9.8     &  6.9   &  14.9   &   8.9     &     19.0         \\
3DAdvDiff + EAG     &  99.9        &    \textbf{70.8}        & \textbf{\textit{99.9}}      &  \textbf{79.5}         &  \textbf{75.9}        &  \textbf{45.3}  & \textbf{\textit{99.9}}   &  \textbf{54.3}      &   \textbf{65.2}     \\ 

\Xhline{3\arrayrulewidth}       
\end{tabular}
}
\end{center}
   
\vspace{-0.2in}
\label{tab:ag}
\end{table}

\noindent \textbf{Generation Quality Augmentation}. Current 3D distance measurements take into account the difference between the entire point set. Therefore, to improve the generation quality, we should limit the $\ell_0$ distance between the adversarial and benign examples. The proposed augmentation notably enhances the quality of the generated point clouds without compromising the attack performance. The results are given in Figure \ref{fig:ab}.

\section{Discussion}

Experiments demonstrate that current attacks perform poorly against black-box models under the $\ell_\text{inf}=0.16$ constraint, particularly in the Chair, Airplane, and Car categories. However, these black-box models are extremely vulnerable to the proposed 3DAdvDiff due to the long-tail training dataset. Consequently, we advocate for a more balanced training approach for 3D point cloud models and the creation of large-scale datasets with a similar scale to the 2D ImageNet. While 3DAdvDiff delivers satisfactory attack performance, its major weakness lies in the need for improved time efficiency to ensure better generalization.

\section{Conclusion}

In this paper, we introduce the first-ever method designed to execute a black-box adversarial attack on recently developed 3D point cloud classification models. Our research is also a pioneering work in the use of diffusion models for 3D adversarial attacks. Specifically, we generate adversarial examples through 3D adversarial shape completion, ensuring reliable and high-quality point cloud generation. We propose several strategies to enhance the transferability of our proposed attack, including the use of model uncertainty for improved prediction inference, enhancing adversarial guidance through ensemble logits from various substitute models, and the improvement of generation quality via critical points selection. Comprehensive experiments on the robust dataset validate the effectiveness of our proposed attacks. Our methods establish a solid baseline for future development in black-box 3D adversarial attacks.

\section*{Acknowledgements}
This work was supported in part by HK RGC GRF under Grant PolyU 15201323.
\bibliographystyle{splncs04}
\bibliography{main}
\end{document}

\title{Supplementary Materials for \\
Transferable 3D Adversarial Shape Completion using Diffusion Models} 

\titlerunning{Transferable 3D Adversarial Shape Completion}

\author{Xuelong Dai\inst{1}\orcidlink{0000-0001-6646-6514} \and
Bin Xiao\inst{1}\orcidlink{0000-0003-4223-8220}}

\authorrunning{X.~Dai et al.}

\institute{
The Hong Kong Polytechnic University\\
\email{xuelong.dai@connect.polyu.hk, b.xiao@polyu.edu.hk }}

\maketitle

%%%%%%%%% BODY TEXT

\section{ModelNet40}

\begin{figure}[t]
   \begin{center}
     \includegraphics[width=1.0\linewidth]{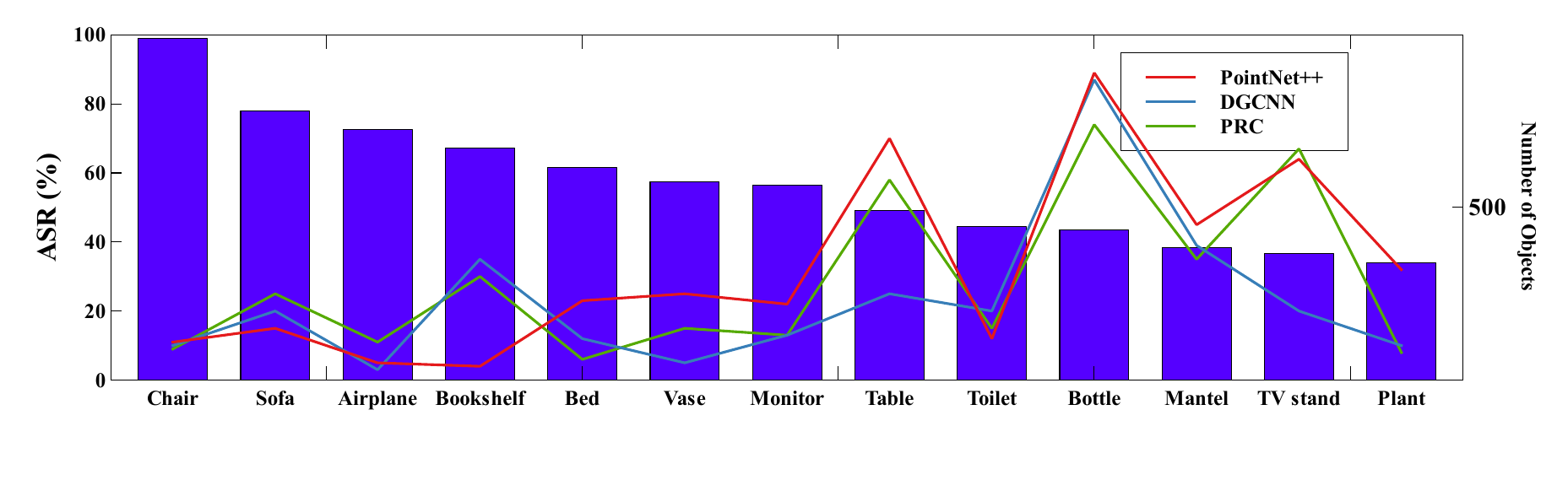}
   \end{center}
      \caption{\textbf{The black-box ASR on the ModelNet40 dataset.} We use the top 13 classes from the ModelNet dataset to demonstrate the long-tailed dataset problem. We use PGD with $\ell_\text{inf}=0.16$ on PointNet to evaluate the black-box attack success rate (ASR).}
   \label{fig:1}
   \end{figure}

We report similar long-tail problems in the ModelNet40 \cite{wu20153d} dataset as in Figure \ref{fig:1}. To further validate the performance of the proposed 3DAdvDiff, we perform experiments on the ModelNet40 dataset. We select the top 8 classes to train the PVD model for shape completion with enough training data. The results are shown in Table \ref{tab:1} and \ref{tab:2}. The proposed 3DAdvDiff$_\text{ens}$ outperforms existing attack methods remarkably on both black-box and against defenses.

Much like the ShapeNet dataset, black-box adversarial attacks typically perform poorly on categories within the ModelNet40 dataset that have a larger volume of training data. However, the test set of ModelNet40 is not uniformly selected. Instead of selecting a fixed proportion from the training data, ModelNet40 chooses 100 point clouds from all the top categories. As a result, the black-box Attack Success Rate (ASR) on ModelNet40 is relatively higher than that on the ShapeNet dataset. However, our proposed 3DAdvDiff$_\text{ens}$ still performs remarkably better than the previous attack methods.

\begin{table}[ht]
\caption{\textbf{The attack success rate (ASR \%) of transfer attack on the ModelNet40 dataset.} The adversarial examples of existing attack methods are generated from the PointNet model. The Average ASR is calculated among the seven black-box models (3DAdvDiff$_\text{ens}$ is calculated among the five black-box models). }

\vspace{-0.2in}
\begin{center}
\resizebox{0.99\columnwidth}{!}{  
\begin{tabular}{c|l|c|cccccccc}

\Xhline{3\arrayrulewidth}
Dataset                                     & Method    & PointNet & PointNet++ & DGCNN & PointConv & CurveNet & PCT & PRC & GDANet & Average \\ \hline
\multicolumn{1}{c|}{\multirow{7}{*}{Chair}} & PGD       &   99.9       &    11.6        &   9.2    &   27.5        &    4.5      &  15.2   &   7.1  &   9.0     &   12.0      \\
 & KNN       &   99.9       &    11.2        &  10.5    &   14.2        &    4.5      &  8.9   &  6.7  &   9.2     &   9.3      \\
& GeoA3     &  99.9        &    7.1       & 4.6      &  8.1         &  3.5        &  5.2   &  6.5   & 4.9      &   5.7      \\
& SI-Adv    &  99.9        &   65.4         &  37.4     &   30.4       &  20.2        &   13.8  &   22.8 &    19.4    &    29.9     \\ \cline{2-11} 
& AdvPC     &   99.9       &     5.1       &   2.6    &   17.2        &    2.4      &  6.2   &   4.2  &   6.8     &    6.4     \\
& PF-Attack &    96.8      &    17.5       &   19.7     &   42.4     &    15.2   &  10.1  &  8.9   &   16.0      &    18.5     \\
& 3DAdvDiff    &    99.9      &    85.4        &  60.8     &   70.6        &    32.1    &  40.8   &  50.6   &  38.9     &     54.2    \\
& 3DAdvDiff$_\text{ens}$    &   99.9       &    \textbf{99.8}        &    \textbf{\textit{99.9}}   &     \textbf{99.5}      &    \textbf{98.7}      & \textbf{90.4}   &  \textbf{\textit{99.9}}    &  \textbf{94.8}      &    \textbf{96.6}     \\ 
\Xhline{3\arrayrulewidth}
\multicolumn{1}{c|}{Dataset}                & Method    & PointNet & PointNet++ & DGCNN & PointConv & CurveNet & PCT & PRC & GDANet & Average \\ \hline
\multicolumn{1}{c|}{\multirow{7}{*}{All}}   & PGD       &   99.9       &   22.8        &   18.8    &   54.5       &   18.3      &  16.1  &   15.4  &   23.4   &   24.1     \\
 & KNN       &   99.9       &   29.6        &   23.4    &   27.5        &    22.3      &  25.9   &  26.5  &   24.5     &   25.7      \\
& GeoA3     &  99.8        &    15.3        & 9.5      &  15.6         &  10.3        &  10.6  &  12.4   &  10.8      &   12.1      \\
& SI-Adv    &  99.9        &  60.4        &  30.1     &    58.4       &  34.8        &   36.8  &24.5  &    36.8    &    40.2     \\ \cline{2-11} 
& AdvPC     &   99.9       &     12.4      &   9.3    &   23.4        &    9.5      &  10.1   &   8.3  &   10.8     &    12.0     \\
& PF-Attack &    99.1      &    51.4        &  31.3     &   67.4        &   35.4       &   35.4  &  23.1   &   41.2    &    40.7     \\
& 3DAdvDiff    &    99.9      &    90.1        &   65.8     &   85.4        &    52.8    &  63.9   &  51.8   &   67.4     &    68.2   \\
& 3DAdvDiff$_\text{ens}$    &      99.9    &    \textbf{99.8}        &   \textbf{\textit{99.9}}     &    \textbf{99.9}       &  \textbf{98.6}        & \textbf{91.2}   &  \textbf{\textit{99.9}}    &  \textbf{96.3}      &    \textbf{97.2} \\ 
\Xhline{3\arrayrulewidth}
\end{tabular}
}
\end{center}
   
\vspace{-0.2in}
\label{tab:1}
\end{table}

\begin{table}[ht]
\caption{\textbf{The attack success rate (ASR \%) of different adversarial attack methods against defenses.} All attacks are evaluated under white-box settings against the PointNet model.}

\vspace{-0.2in}
\begin{center}
%\resizebox{1.0\columnwidth}{!}{  
\begin{tabular}{l|c|ccccc}

\Xhline{3\arrayrulewidth}
Method    & ASR & SRS & SOR & DUP-Net & IF-Defense & HybridTraining \\ \hline
PGD       & 99.9 &  61.3   & 17.6  &   16.5      &     14.3       &    0.4           \\
KNN       & 99.9 &  94.5   & 85.4  &   48.9     &     22.4       &    12.1    \\
GeoA3     & 99.8 & 55.3  &  28.6  &      22.1     &      13.6      &  1.5             \\
SI-Adv    & 92.5 &  75.1   & 22.1    &   20.3    &  18.6        &     19.1           \\ \hline
AdvPC     & 99.9 & 84.8    &  21.4   &  19.8      &    20.6       &      0.4          \\
PF-Attack & 99.1 &  47.5   &  77.3   &   43.0     &   29.2        &    13.6            \\
3DAdvDiff    & 99.9    & 95.6    &   90.5      &     88.3       &    \textbf{52.1}     &   31.5   \\
3DAdvDiff$_\text{ens}$    & \textbf{99.9} &  \textbf{98.7}   &  \textbf{96.0}   &    \textbf{95.4}     &   43.7         &  \textbf{98.6}     \\ 
\Xhline{3\arrayrulewidth}        
\end{tabular}
%}
\end{center}
   
\vspace{-0.2in}
\label{tab:2}
\end{table}

\section{Acceleration}
The original PVD model adopts the DDPM \cite{ho2020denoising} sampling for generating point clouds, which use 1000 sampling steps. An effective acceleration method to improve the sampling of DDPM is to use DDIM \cite{song2020denoising} sampling. We implement DDIM sampling for the PVD model with only 200 sampling steps. The results are shown in Table \ref{tab:3}. We significantly improve the sampling speed without largely decreasing the generation quality. Improving time efficiency is a hot topic in the community, with many acceleration methods being introduced. Therefore, we believe the time efficiency of diffusion model based adversarial attacks can be further enhanced in the future.

\begin{table}[t]
\caption{\textbf{The attack performance with DDPM and DDIM sampler.} All attacks are evaluated under white-box settings against the PointNet model on all selected classes of the ShapeNet dataset.}

\vspace{-0.2in}
\begin{center}
%\resizebox{1.0\columnwidth}{!}{  
\begin{tabular}{l|ccc}

\Xhline{3\arrayrulewidth}
     & ASR  & Time(s) & CD   \\ \hline
DDPM & \textbf{90.1} & 60.8    & \textbf{0.14} \\
DDIM & 89.9 & \textbf{13.5}    & 0.18 \\

\Xhline{3\arrayrulewidth}
\end{tabular}
%}
\end{center}
   
\vspace{-0.2in}
\label{tab:3}
\end{table}

\section{Visual Results}
\begin{figure}[ht]
   \begin{center}
     \includegraphics[width=1.0\linewidth]{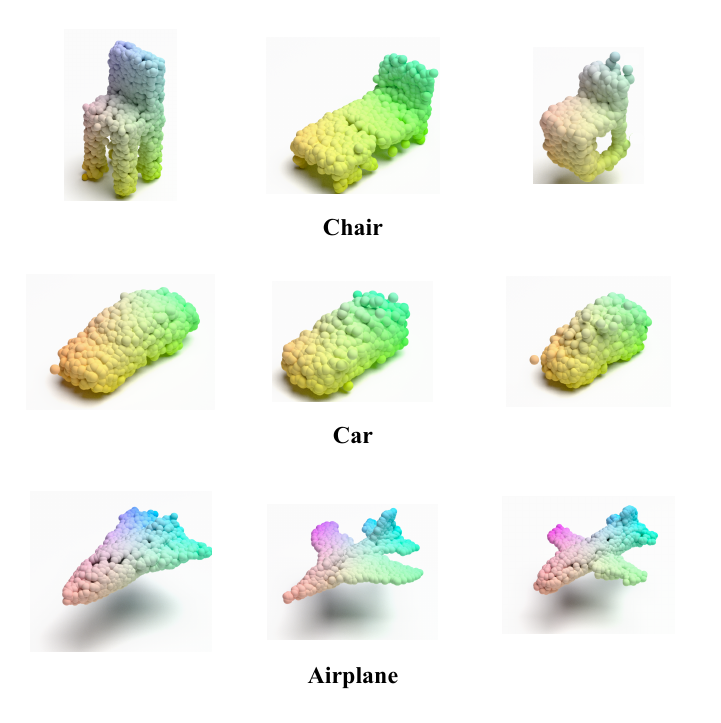}
   \end{center}
      \caption{\textbf{The generated adversarial point clouds of 3DAdvDiff$_\text{ens}$.}}
   \label{fig:2}
   \end{figure}

\begin{figure}[ht]
   \begin{center}
     \includegraphics[width=1.0\linewidth]{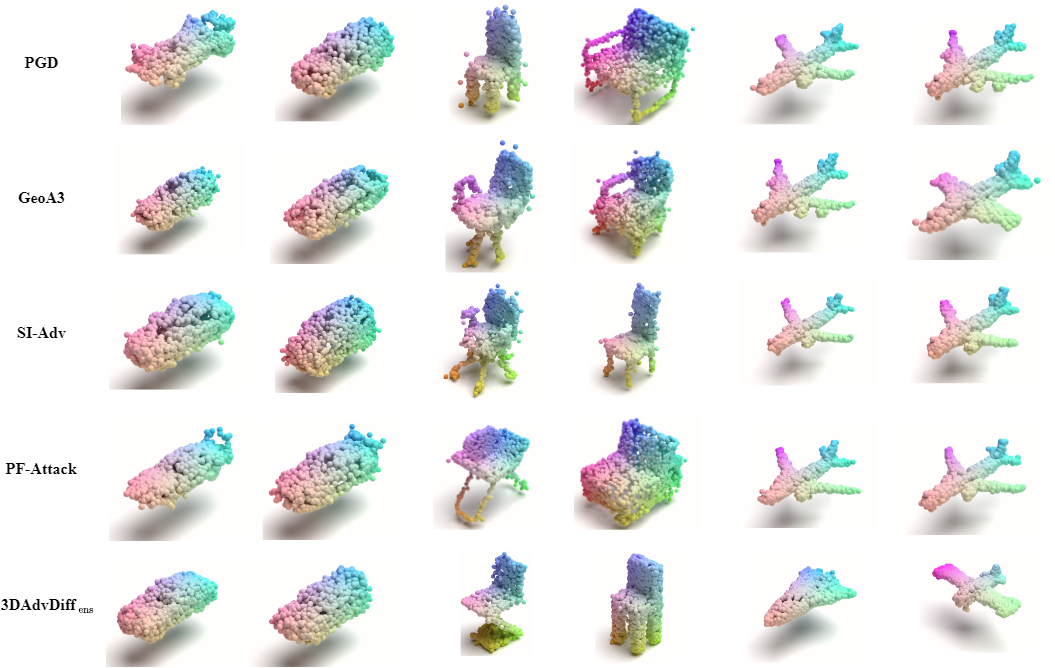}
   \end{center}
      \caption{\textbf{The generated adversarial point clouds.} The adversarial examples are randomly sampled from ShapeNet dataset.}
   \label{fig:3}
   \end{figure}

We further give the visual results of our generated 3D adversarial point clouds in Figure \ref{fig:2} and \ref{fig:3}.

\section{Multi-View Adversarial Shape Completion}

The shape completion tasks performed by the PVD model generate 20 different views for a specified partial shape to generate 20 different point clouds, which enables us to locate the most vulnerable views for generating adversarial point clouds. A similar finding is also addressed by Zhao et al. \cite{zhao2020isometry}. Therefore, it further enhances the performance of the proposed attacks as 3D deep learning models are sensitive to the transformations of 3D point clouds.

\section{Adversarial Shape Generation}

Diffusion models also have the ability to directly generate complete 3D point clouds without the need for a given partial shape. We further evaluated the performance of our proposed 3DAdvDiff in the context of adversarial shape generation, which we refer to as 3DAdvDiff-Gen. As demonstrated in Table \ref{tab:4} and \ref{tab:5}, the 3DAdvDiff model, when used for adversarial shape generation, outperforms shape completion in black-box attacks, showing an average increase of 7.8\% in Attack Success Rate (ASR). However, since shape generation does not inherently support multi-view generation during its original training, the white-box ASR is somewhat compromised without identifying the vulnerable transformations. Despite this, both time efficiency and training efficiency are enhanced. It should be noted, however, that the quality of adversarial shape generation is somewhat worse compared to shape completion. This could potentially be due to the absence of guidance on partial shape. We leave a better design of the  shape generation in future works.

\begin{table}[ht]
\caption{\textbf{The attack success rate (ASR \%) of adversarial shape completion and shape generation.} The adversarial examples of existing attack methods are generated from the PointNet model on the ShapeNet's Chair class. The Average ASR is calculated among the seven black-box models.}

\vspace{-0.2in}
\begin{center}
\resizebox{0.99\columnwidth}{!}{  
\begin{tabular}{l|c|cccccccc}

\Xhline{3\arrayrulewidth}
 Method    & PointNet & PointNet++ & DGCNN & PointConv & CurveNet & PCT & PRC & GDANet & Average \\ \hline
 3DAdvDiff   & \textbf{99.9}      &    \textbf{60.6}        &   8.7     &   23.5        &     9.8     &  6.9   &  14.9   &   8.9     &     19.0   \\
 3DAdvDiff-Gen    &   90.1  &  54.2     &   \textbf{25.4}        &    \textbf{32.1}  &     \textbf{21.2}     &    \textbf{7.6}      &  \textbf{24.5}   &  \textbf{22.5}    &  \textbf{26.8} \\  
\Xhline{3\arrayrulewidth}
\end{tabular}
}
\end{center}
   
\vspace{-0.2in}
\label{tab:4}
\end{table}

\begin{table}[ht]
\caption{\textbf{The generation quality on the ShapeNet dataset.} The CD distance is multiplied by 10$^{-2}$. }

\vspace{-0.2in}
\begin{center}
%\resizebox{1.0\columnwidth}{!}{  

\begin{tabular}{l|cc}

\Xhline{3\arrayrulewidth}
Method  & 3DAdvDiff$_\text{ens}$ & 3DAdvDiff$_\text{ens}$-Gen \\ \hline
HD     &          \textbf{0.098}               &     0.80                  \\
CD     &         \textbf{0.14}                &    0.36                   \\
MSE     &       \textbf{1.18}                  &     3.05                
\\ \Xhline{3\arrayrulewidth} 
\end{tabular}
%}
\end{center}
   
\vspace{-0.2in}
\label{tab:5}
\end{table}

\section{Limitation}

Given the unique characteristics of 3D point clouds, they necessitate a larger volume of training data compared to 2D images when training diffusion models. At present, all existing 3D diffusion models are trained using the large-scale classes in the ShapeNet dataset. This, however, restricts the generalizability of the proposed diffusion adversarial attacks to relatively smaller datasets. Nonetheless, we are optimistic that with the continued advancement of 3D diffusion models, a large-scale and balanced 3D dataset will become available in the future. Furthermore, while we have managed to enhance the sampling speed of our proposed 3DAdvDiff with DDIM sampling, the generation speed of the proposed attack still lags behind PGD-based attacks. However, with the rapid development of diffusion models, the time efficiency problem is addressed in many recent works. Our future goal is to further boost the efficiency of 3DAdvDiff by incorporating acceleration techniques from diffusion models.

\section{Ethics Concerns}

The proposed 3DAdvDiff brings new challenges to 3D deep learning models. Adversaries may adopt our attacks to generate malicious point clouds to attack the 3D deep learning classification models. However, our proposed 3DAdvDiff can also be utilized for adversarial training to enhance the robustness of 3D deep learning models. The proposed attack can further encourage the development of 3D adversarial defenses. Therefore, our proposed 3DAdvDiff can achieve positive impacts on improving the 3D deep learning model robustness.

\bibliographystyle{splncs04}
\bibliography{main}